\documentclass[sn-chicago,Numbered]{sn-jnl}% Chicago-based Humanities Reference Style
\usepackage{graphicx}%
\usepackage{multirow}%
\usepackage{amsmath,amssymb,amsfonts}%
\usepackage{amsthm}%
\usepackage{mathrsfs}%
\usepackage[title]{appendix}%
\usepackage{xcolor}%
\usepackage{textcomp}%
\usepackage{manyfoot}%
\usepackage{booktabs}%
\usepackage{algorithm}%
\usepackage{algorithmicx}%
\usepackage{algpseudocode}%
\usepackage{listings}%
\usepackage{bm}
\usepackage{hyperref}
\usepackage{textcomp}
\usepackage{amsmath}
\usepackage{bbm}
\theoremstyle{thmstyleone}%
\theoremstyle{thmstyletwo}%

\theoremstyle{thmstylethree}%

\begin{document}

\title[Article Title]{Classification of Breast Cancer Histopathology
Images using a Modified Supervised Contrastive Learning Method}

\author[1]{\fnm{Matina} \sur{Mahdizadeh Sani}}\email{matinamahdizadehsani@gmail.com}

\author[2]{\fnm{Ali} \sur{Royat}}\email{aliroyat73@gmail.com }

\author*[1]{\fnm{Mahdieh} \sur{Soleymani Baghshah}}\email{soleymani@sharif.edu}

\affil*[1]{\orgdiv{Computer Science and Engineering}, \orgname{Sharif University of Technology}, \orgaddress{\city{Tehran}, \country{Iran}}}

\affil[2]{\orgdiv{Electrical Engineering}, \orgname{Sharif University of Technology}, \orgaddress{\city{Tehran}, \country{Iran}}}


\abstract{Deep neural networks have reached remarkable achievements in medical image processing tasks, specifically in classifying and detecting various diseases. However, when confronted with limited data, these networks face a critical vulnerability, often succumbing to overfitting by excessively memorizing the limited information available. \textcolor{black} {This work addresses the challenge mentioned above by improving the supervised contrastive learning method leveraging both image-level labels and domain-specific augmentations to enhance model robustness. This approach integrates self-supervised pre-training with a two-stage supervised contrastive learning strategy. In the first stage, we employ a modified supervised contrastive loss that not only focuses on reducing false negatives but also introduces an elimination effect to address false positives. In the second stage, a relaxing mechanism is introduced that refines positive and negative pairs based on similarity, ensuring that only relevant image representations are aligned.} We evaluate our method on the BreakHis dataset, which consists of breast cancer histopathology images, and demonstrate an increase in classification accuracy by 1.45\% in the image level, compared to the state-of-the-art method. This improvement corresponds to 93.63\% absolute accuracy, highlighting the effectiveness of our approach in leveraging properties of data to learn more appropriate representation space. The code implementation of this study is accessible on GitHub\footnote{\href{https://github.com/matinamehdizadeh/Breast-Cancer-Detection}{https://github.com/matinamehdizadeh/Breast-Cancer-Detection}}.
}


\maketitle

\section{Introduction}\label{sec1}

Cancer, a leading cause of mortality worldwide, poses a significant threat to public health. Among the various types of cancer, breast cancer is the most prevalent form affecting women, resulting from abnormal cell division in the breast tissue. However, it is essential to note that not all tumors are life-threatening, as they can be either benign or malignant, and 
detecting a tumour's malignancy or benign nature is crucial in determining appropriate treatment strategies and ensuring patient well-being.

Medical imaging stands out as one of the most commonly used applications of artificial intelligence in healthcare. Resent research focused on applying machine learning to medical images demonstrates significant potential in revolutionizing the diagnostic process of various cancerous diseases due to enabling the early detection and accurate characterization of tumors \cite{ma2023implementation, an2023comprehensive}. However, several challenges within the medical field are delaying the widespread utilization of machine-learning capabilities in real-world applications. These challenges include the generalization issues of Deep Neural Networks, collecting diverse data sets, acquiring sufficient labeled data for training, and the risks and high costs associated with errors in medical decision-making. The growing importance of these methods in the medical domain highlights the urgent need to implement effective approaches to overcome these challenges more than ever.

The classification of medical images plays a supportive role in clinical treatment. Histopathology also greatly benefits from the application of machine learning techniques. Leveraging the power of machine learning in diagnosing cancerous images has the potential to revolutionize the field, aiding pathologists to make precise diagnoses and facilitating the treatment process. However, due to the critical nature of this issue and the potential risks, the widespread use of these methods is still limited. As a result, it is crucial to focus on reducing existing errors by examining earlier methods and striving to improve upon them. Supervised methods previously used for image classification have their limitations, as they require a considerable amount of labeled data and collecting them usually needs high cost. Fortunately, in recent years, self-supervised learning has emerged as a promising approach, presenting its effectiveness in various tasks.

The contributions of this work are focused on overcoming the challenges associated with training networks in the context of cancer image diagnosis. One of the main challenges addressed in this study is the scarcity of labeled data in the medical domain, which poses difficulties for supervised learning approaches. To tackle this issue, the proposed method draws inspiration from self-supervised learning techniques and extends them by utilizing labeled data and leveraging similarities between images in the representation space. Additionally, to address the problem of generalization in deep neural networks, various data augmentation techniques specific to histopathology datasets and H\&E stains are employed. 
\textcolor{black}{In the process of diagnosing a carcinogenic tissue sample, pathologists analyze specific features of histopathological images. These features include the morphological characteristics of cellular structures, such as the size and shape of nuclei, the density of cells, and the arrangement of tissues. The color of the sample, influenced by H\&E stain, also plays a crucial role in highlighting these features. Hematoxylin stains nuclei blue, while Eosin stains extracellular structures pink, creating a contrast that is essential for differentiating between benign and malignant tissues. However, the consistency of these stains can vary due to differences in reagent concentrations, exposure to light, and tissue scanning techniques, which may lead to color variations across different laboratories. Hence, the augmentation techniques help to enhance the robustness and generalization capabilities of the trained models \cite{marini2023data}.}

The primary objective of this study is to develop a method that improves the accuracy of cancer image diagnosis in two publicly available breast cancer datasets. The proposed method considers all available magnifications, ensuring comprehensive and accurate classification of the images. The method's effectiveness is evaluated by comparing it to previous works, and a detailed analysis of the obtained results is presented at the end.

\section{Literature Review}
\label{sec:literature}
Over the past few years, machine learning has achieved remarkable progress in developing systems by leveraging large amounts of labeled data. While this learning approach has shown excellent performance in previous tasks, it faces limitations in medical applications due to the issue of data scarcity.

Self-supervised learning is an approach that does not need labeled data to train the deep models. It can be divided into two main subcategories: contrastive and non-contrastive. Contrastive learning aims to maximize the similarity between positive pairs and minimize the similarity between negative pairs, e.g., SimCLR \cite{chen2020simple}, SimCLRv2 \cite{chen2020big}, and PIRL \cite{misra2020self} are based on this approach. On the other hand, non-contrastive learning methods works in other ways, for example, they focus only on maximizing the similarity between positive pairs since finding negative pairs can be challenging when working with unlabeled data. Some examples of non-contrastive learning e.g. BYOL \cite{grill2020bootstrap}, DirectPred \cite{tian2021understanding}, and SimSiam \cite{chen2021exploring}. Furthermore, a more effective approach to contrastive learning is to utilize the labels associated with the data to guide the choice of negative and positive pairs, where positive pairs consist of samples that belong to the same class and negative pairs are formed by taking samples from different classes \cite{khosla2020supervised}.

Results of research conducted on medical images using self-supervised methods demonstrate better performance compared to the supervised ones. For instance, 
\textcolor{black}{Spathis et al. \cite{spathis2022breaking}, using electrocardiogram data, illustrated that clever transformations of input data with self-supervised approaches can perform well even without extensive annotations. Peng et al. \cite{peng2024boundary} used Information Invariant Clustering to develop local representations without relying on traditional contrastive learning, advancing self-supervised learning for medical image segmentation tasks. Their method introduces a regularized mutual information objective and a boundary-aware loss, enhancing model interpretability.} Chhipa et al. \cite{chhipa2023magnification} introduced a method based on the Efficient-net \cite{tan2019efficientnet} architecture that utilizes the SimCLR approach for the binary classification of the BreakHis dataset. Jin et al. \cite{math11010110} introduced the HistoSSL architecture, a knowledge distillation model based on non-contrastive learning, which does not rely on a strategy for creating negative pairs. Additionally, Li et al. \cite{li2021dual} proposed a combined method of SimCLR and multiple instance learning \cite{dietterich1997solving}, in which the representations of an image are extracted through self-supervised learning and given as input to a multiple instance learning model. Ciga et al. \cite{ciga2022self} presented a relatively comprehensive article using the self-supervised SimCLR method on different ResNet \cite{he2016deep} architectures, i.e. ResNet-18, ResNet-34, ResNet-50, and ResNet-101.
The network was trained in a self-supervised manner on 57 different histopathology datasets, and ultimately, the weights obtained were used for classification, segmentation, and regression tasks on the data. \textcolor{black}{Additionally, Ghesu et al. \cite{ghesu2201self} demonstrated the utility of contrastive learning and online feature clustering on 100 million medical images from diverse modalities, validating their approach on multiple medical problems. Wang et al. \cite{wang2023pyramid} proposed the Pyramid-based Local Wavelet Transformer model, which enhances self-supervised learning by integrating a pyramid-based transformer into the encoder of Masked Autoencoders (MAE). This approach is designed to effectively capture both local and global features in histopathological images, improving the representation learning process in self-supervised settings.}

The cases mentioned above represent only a few examples of the research conducted in the field of histopathology for self-supervised learning. There is comparatively less research in this area than in supervised learning,  creating an opportunity for progress and advancement, as demonstrated in this report.

\section{Method}\label{sec3}
This work mainly focuses on leveraging the advantages of self-supervised learning to extract high-quality representations from images. However, as self-supervised learning may produce false negatives and positives when creating the pairs, we complement it with supervised learning, which uses image labels, to mitigate these effects.

\begin{figure}[!t]
\centering
\centerline{\includegraphics[width=\columnwidth]{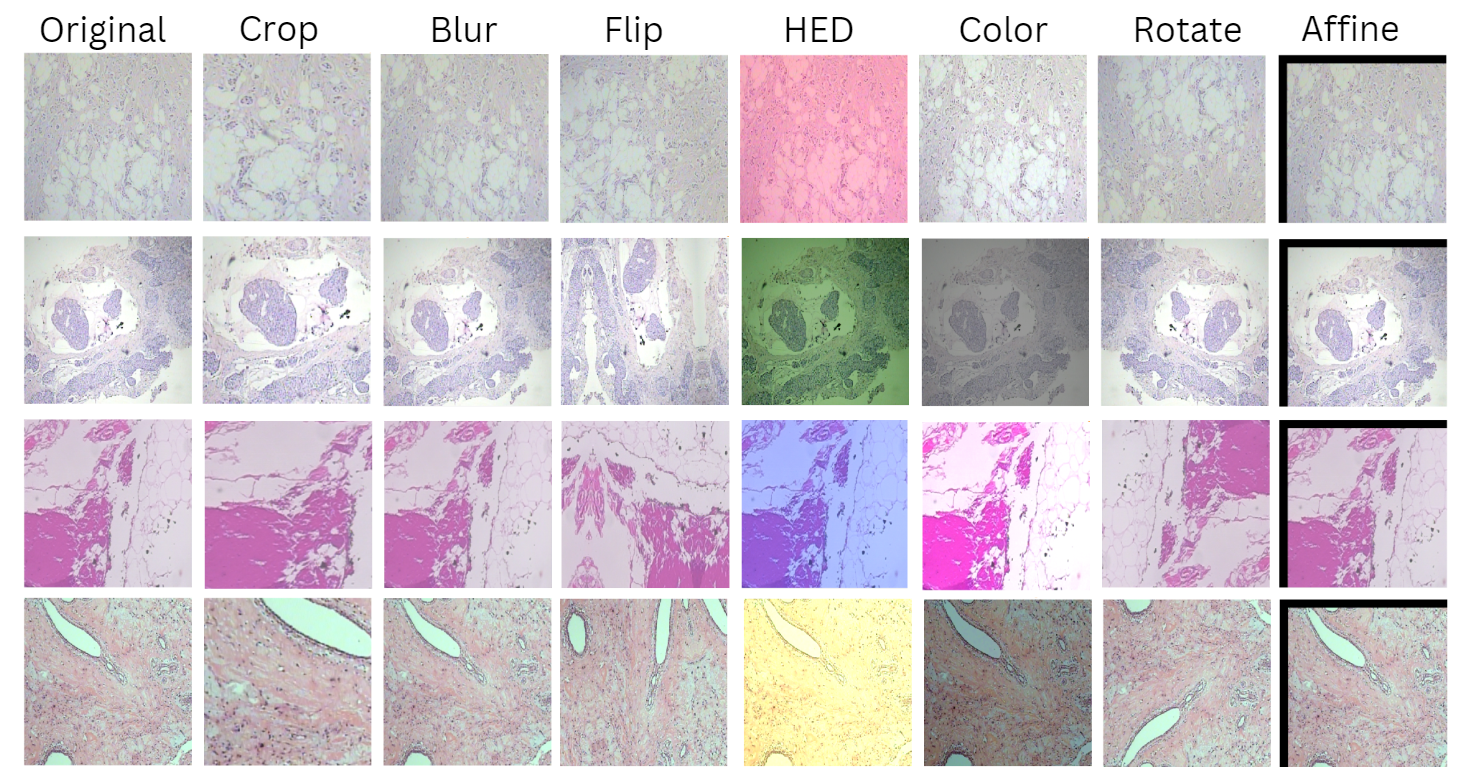}}
\caption{Data augmentation strategies. In this figure, an illustration of the various data augmentation methods used in the research is provided.}\label{augs}
\end{figure}

First, the model is trained on five distinct breast cancer datasets, with the SimCLR technique to establish a solid starting point. We then utilize these pre-trained weights as the initial weights for our representation learning phase. The representation learning phase includes a two-stage supervised contrastive learning. Following this, supervised fine-tuning is accomplished
wherein three fully connected layers are added to the end of the encoder for classification task and one fully connected layer is employed for an auxiliary task. Ultimately, four identical architectures are employed for four magnification factors in the dataset and trained separately.

%\subsection{Pair Making Strategies}
\subsection{Augmentations}
The initial step in using a self-supervised approach is to generate positive and negative pairs. In this study, two techniques were employed to create such pairs.

\begin{figure}[!t]
\centerline{\includegraphics[width=0.97\textwidth]{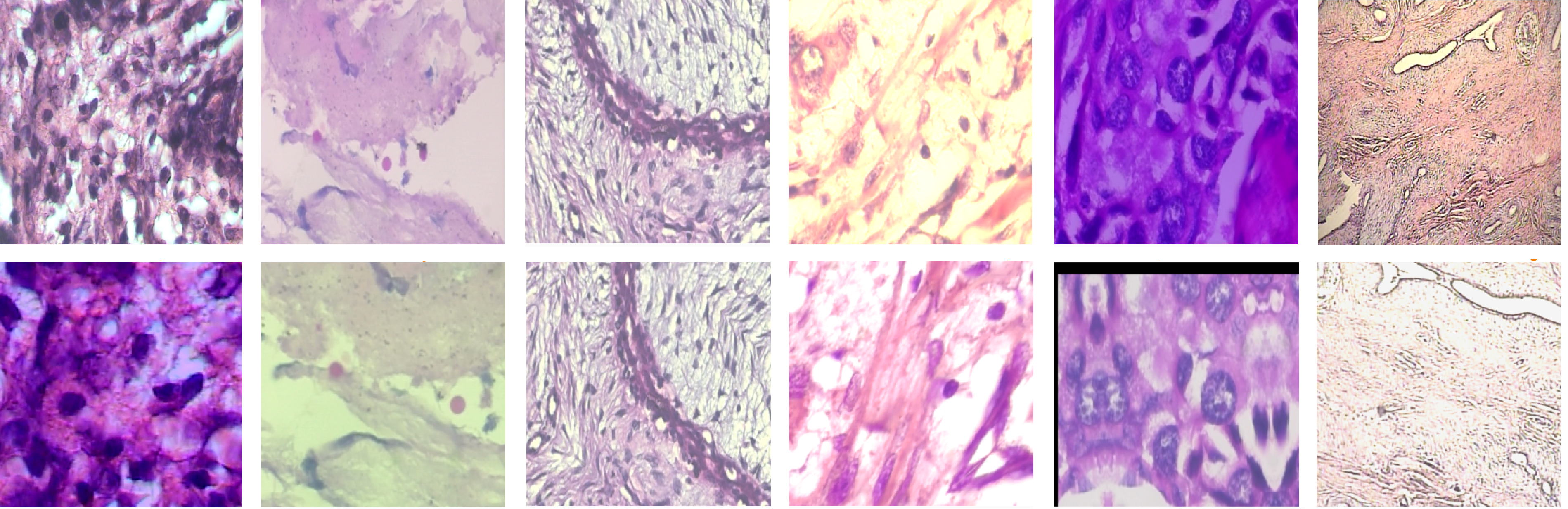}}
\caption{Positive pairs. This figure depicts an anchor image alongside one of its corresponding positive pairs. This pair is created through a combination of data augmentations introduced in figure \ref{augs}, and each augmentation is applied with a specific probability. The first row indicates the anchor image, while the second one is donated to the corresponding positive pair.
}\label{final-augs}
\end{figure}

The most prevalent way of creating positive pairs is through augmentation. In this study, various methods were utilized for this purpose, including Random Crop, Color Jitter, Gaussian Blur, and Geometric %Augmentationall, of
which are depicted in Figure \ref{augs}. All of the mentioned augmentations are applied with a fixed probability to create the image pairs, and the resulting pairs can be observed in Figure \ref{final-augs}.

Furthermore, a histopathology-specific augmentation technique known as HED \cite{tellez2019quantifying} has been employed. This method involves separating the two H\&E stains from the RGB images, adjusting the brightness and contrast of each stain's corresponding matrix individually, and then converting them from the H\&E channel to the RGB. This process serves as a data augmentation method for creating positive pairs and also enhances the model's ability to generalize against different stains. These stains can be separated using the following formula \cite{yang2021self}.

\begin{equation}
    V = log \frac{I_0}{I} = WH\label{hed1}
\end{equation}
According to Beer-lambert law $I \in R^{3 \times n}$ is the RGB density matrix, and $I_0$ is equal to 255. $V \in R^{3 \times n}$ is an optical density matrix, $W \in R^{3 \times 2}$ is the conversion matrix, and $H \in R^{2 \times n}$ is the stain matrix\cite{vahadane2016structure}.
\textcolor{black}{}{The following optimization problem is utilized to find desired factorization:}
\begin{equation}
\begin{split}
\min_{W,H} \frac{1}{2}||V - WH||^2_F + \lambda \sum^{2}_{j=1} ||H(j,:)||_1,
\\
s.t\: W,H \ge 0, ||W(:,j)||^2_2 = 1
\end{split}
\end{equation}\label{hed2}

By solving this sparse non-negative matrix factorization problem mentioned above, $W$ and $H$ can be estimated \cite{yang2021self}.
$H$ is the converted matrix, from RGB image to $H$ and $E$ arrays. Subsequently, in HED augmentation, the contrast and brightness of $I_h$, and $I_e$ are altered as:
\begin{equation}
I_h = aug(I_0 exp(-H[0,:])), I_e = aug(I_0 exp(-H[1,:]))
\end{equation}\
Finally, stains are converted back into the RGB color space using the $W$ matrix \cite{yang2021self}.

\subsection{Contrastive Representation Learning}
\begin{figure*}[!t]
\centerline{\includegraphics[width=1\textwidth]{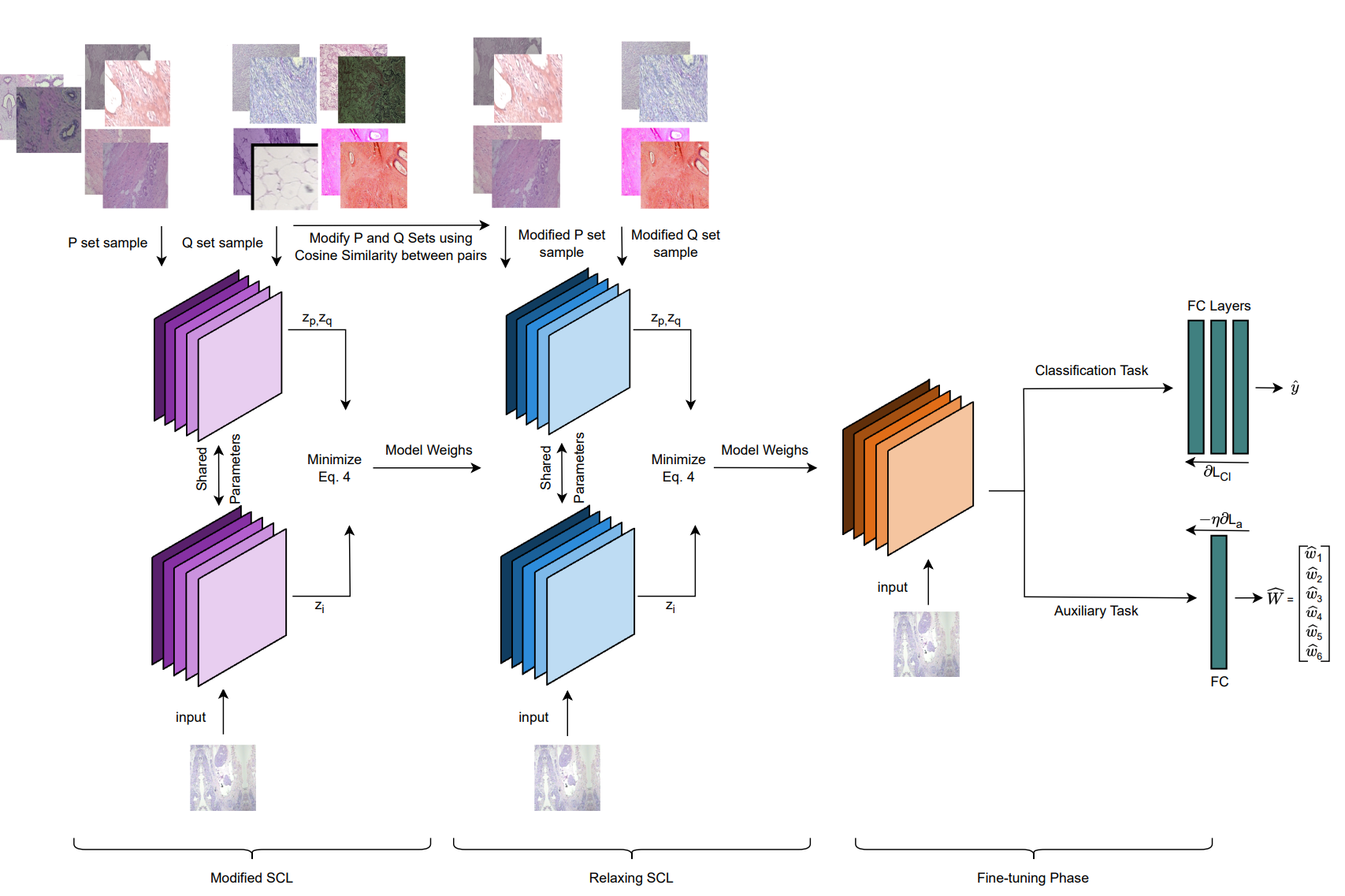}}
\caption{Network architecture. The architecture consists of three training stages, starting with the weights of EfficientNet pre-trained on histopathology images. The pre-training phase involves training on the Breakhis dataset using the modified supervised contrastive learning method. Subsequently, the similarity between the image representations is calculated to correct the positive pairs, followed by a retraining with the updated set of pairs in the relaxing phase. The final stage is the fine-tuning witch involves an auxiliary and classification task using a supervised approach. Here, the representation of the anchor image is denoted as $z_i$, while $z_p$, and $z_q$ are the representation of a positive pair and a negative pair 
respectively. $\hat{y}$ and $\hat{W} = (\hat{w_1}, \hat{w_2}, \hat{w_3}, \hat{w_4}, \hat{w_5}, \hat{w_6})$ are the predicted label, and conversion matrix with respect to the input image.}\label{arch}
\end{figure*}

In this phase, we have two stages of training with the contrastive method, as illustrated in Figure \ref{arch}. 

\subsubsection{Modified Supervised Contrastive Learning}
First, we extract the representation using a modified supervised contrastive loss which takes the following form.

\begin{equation}
 \ell_{i} = \frac{- \alpha}{|P(i)|} \sum_{p\in P(i)}
 \log( \frac{exp(z_i \cdot z_p/\tau)}{\sum_{p\in P(i)} exp(z_i \cdot z_p/\tau) + \lambda \sum_{q\in Q(i)} exp(z_i \cdot z_q/\tau)})
\label{floss-def}
\end{equation}

During the first stage, positive pairs are created using augmentations and image labels. In this context, $i$ is the index of the anchor image. Assuming a set $A$ comprising $2N-1$ samples, from both images and their corresponding augmented pairs, derived from a batch of size $N$, we define $P$ as the subset of images within set $A$ sharing the same label as the anchor, and $Q$ as the subset containing images with different labels from the anchor. The representation of anchor image is denoted as $z_i$, while $z_p$ represents one of the positive pairs for $z_i$ in the $P$ set. Similarly, $z_q$ denotes a negative pair for $z_i$ in the $Q$ set. Furthermore, $\alpha$ is the weight assigned to this loss function, which is calculated with respect to the number of samples in both classes to overcome the class imbalance issue.
The utilization of the dot product instead of similarity stems from the fact that the data is normalized.

\subsubsection{Relaxing Supervised Contrastive Learning}
Once the model has been trained using this loss function, its weights are loaded to calculate the cosine similarity between all pairs of images in the dataset.

In the relaxing stage, the positive pairs are created the same as in the previous phase. However, the set $P$ is altered in such a way that if an image in the $P$ set has a similarity score less than 0.5 in the obtained representation space, it is removed from the $P$ set and is no longer considered as a positive pair. This means that if, after the model is trained, assuming the $i$-th image is a positive pair for the anchor, which shows negative similarity in the resulted representation space, then it is not a good positive pair. More precisely, it does not have enough similarity even after enforcing it to be similar to the anchor, and so it is removed from the $P$ set. The same technique goes for the $Q$ set, which is if an image in the $Q$ set has a similarity score more than 0.5 in the obtained representation space, it is removed from the $Q$ set and is no longer considered as a negative pair.

In this stage, we continue to use the same loss funcion as Eq.~\ref{floss-def}. However, in contrast to the first stage, $z_p$ now represents a positive pair in the $P$ set, and greater than $0.5$ similarity with the anchor. Similarly, $z_q$ denotes as a negative pair for $z_i$ in the $Q$ set, and less than $0.5$ similarity with the anchor. The proposed contrastive learning method is summarized briefly in Algorithm \ref{algo}.

\subsection{Fine-tuning}
In this stage, the weights generated during the relaxing phase are fine-tuned via a supervised process using the labeled data. This supervised process comprises a primary classification task and an auxiliary task.

\subsubsection{Classifier}
 This architecture, depicted in Figure \ref{arch}, is designed to perform binary image classification. To accomplish this task, a classification head consist of three fully connected layers, are added at the end of the encoder. 
This loss function is computed using the following equation.

\begin{equation}
L_{Cl} = \frac{1}{N} \sum_{i=1}^{N} \mathcal{L}_{CE}(\hat{y_i}, y_i) \label{lcl-def}
\end{equation}

$y_i$ is the %label with respect to
ground truth label for $x_i$, and $\hat{y_i}$ is the network's output for the classification task. $\mathcal{L}_{CE}$ is the Cross-Entropy classification loss.

\subsubsection{Auxiliary Task}
An auxiliary task is also incorporated to improve the network's robustness against different H\&E stains by adding one fully 
connected layer at the end of the encoder \cite{man2020classification}.

\begin{algorithm}[t!]
\caption{Improved Supervised Contrastive Learning}\label{algo}
\textbf{Input:} Efficient-net2 model, training dataset $X_{train}$

\textbf{Output:} Efficient-net2 weighs $W$

\begin{algorithmic}[1]
\State Initialize $P$, and $Q$ sets for each image, $x_i$, using data augmentations and the labels
\State calculate Eq. \ref{floss-def}, which Maximize similarity between $x_i$ and its $P$ set in a batch while minimize similarity between $x_i$ and the batch images with double focus ($\lambda = 2$) on ones from the $Q$ set in the representation space
\State Train for 200 epochs
\State Calculate similarity between the representation of all image pairs
\State If similarity $< 0.5$ remove from $P$ set
\State If similarity $> 0.5$ remove from $Q$ set
\State Repeat step 2, and 3.
\State Return $W$
\end{algorithmic}
\end{algorithm}

The output of this task is a six-element matrix, which predicts the W matrix in Eq. \ref{hed1}. The loss function associated with this branch calculates the difference between the predicted and original matrices:
This loss function is computed using the following equation.

\begin{equation}
L_{a} = \frac{1}{N} \sum_{i=1}^{N} \mathcal{L}_{a}(\hat{W_i}, W_i) \label{lossr-def}
\end{equation}
where $W_i$ is the matrix that transforms the input image from the RGB space to the H\&E space, and $\hat{W_i}$ is the network's output for the auxiliary task. $\mathcal{L}_{a}$ measures the difference between the predicted and original transformation matrices using Squared L2.

The resulting loss value is then fed back to the network with a negative sign to reduce the model's sensitivity to variations in stain colors.
The experiments %on this method 
showed that incorporating the auxiliary task alone does not significantly improve the model's accuracy. However, to leverage the benefits of this technique, images must first be augmented using HED and then provided as inputs to the encoder. Moreover, during the fine-tuning stage, other data augmentation methods are utilized to enhance the model's generalization capabilities further and increase the robustness of the learning process.

\subsubsection{Final Loss}
The final loss function is calculated by combining the classification loss and auxiliary loss with the coefficient $\eta$. The equation is as follows:

\begin{equation}
Loss = L_{Cl} - \eta L_a  \label{loss-def}
\end{equation}

\section{Experiment and Results}\label{sec3}
This section investigates the experimental evaluation of the proposed method on two publicly available datasets and compares its results with those of previous methods.
\subsection{Datasets}
\begin{enumerate}
    \item BreakHis \cite{7312934}. This dataset consists of 7,909 stained microscopic images of breast tumor tissue collected from 82 patients. The images are captured at four different magnification factors, namely, $40X$, $100X$, $200X$, and $400X$.
    The dataset consists of two primary breast tumour categories, classified as benign or malignant. Specifically, there are 2,480 samples from benign patients and 5,429 from malignant patients.
    The BreakHis dataset is imbalanced, with significantly more samples in the malignant class. To mitigate the impact of this issue, a weighted loss function was used during all three training stages.
    \item BACH \cite{Aresta_2019}. This dataset comprises 400 stained microscopic images and 30 whole slide images of breast tumor tissue. The microscopic images have been categorized into normal, benign, in situ carcinoma, and invasive carcinoma, with each category containing 100 images. The purpose of employing the BACH dataset was to evaluate the generalization of our backbone, which was trained using the BreakHis dataset. 
    As the research objective revolves around the binary classification of breast tumors, a few adjustments were made to the labels in the BACH dataset. The normal and benign labels were treated as benign, while the in situ carcinoma and invasive carcinoma labels were considered malignant. Similar technique have been employed in previous studies, such as those cited in \cite{zhu2019breast}, and \cite{patil2019breast}.

\end{enumerate}

To conduct the experiments, the datasets was split into training, test, and validation sets in a $60\%,20\%,20\%$ ratio, respectively, and the experiments were repeated for 5 folds to ensure accurate results.

\subsection{Experimentation Details}
The EfficientNet-B2 \cite{tan2019efficientnet} is an encoding architecture that is utilized in this study. To ensure a solid starting point, we begin by randomly assigning the Efficient-Net2 weights. Subsequently, we train the model using SimCLR on five distinct breast cancer datasets: andrewjanowczyk epi \cite{janowczyk2016deep}, andrewjanowczyk mitosis \cite{janowczyk2016deep}, andrewjanowczyk nuclei \cite{janowczyk2016deep}, Mitos \& Atypia \cite{roux2014mitos}, and IDCGrad \cite{bolhasani2020histopathological}. In order to evaluate the performance of the proposed model for binary classification of the BreakHis dataset, some previous networks were chosen as benchmarks. Then, their performance were assessed by comparing the Image-Level Accuracy (the average accuracy across all images) and the Patient-Level Accuracy (the average accuracy for individual patients) against the performance of the proposed model.

The network is trained for $200$ epochs in the first two stages. During this process, the temperature of $0.01$, the learning rate $10^{-5}$, and a set of augmentations is utilized. Additionally, the Adam optimizer is employed as the optimization function, the batch size is set to $12$, and the $\lambda$ parameter is equal to $2$.

In the next stage of training, i.e. the supervised fine-tuning phase, the four distinct models, each for one magnification factor, are trained for only $20$ epochs. The learning rate is set to $2 \times 10^{-5}$, and augmentations are applied to enhance the network's generalization capabilities. Furthermore, the number of fully connected layers and neurons in each layer are determined through hyperparameter tuning. Additionally, dropout with a probability of $0.5$ is implemented between the fully connected layers, the batch size is 8, and the $\eta$ parameter for the auxiliary task is set to $0.5$.

\begin{table}[t]
\caption{\textcolor{black}{Comparison of image-level accuracy with previous methods for four magnification factors of the BreakHis dataset.}}\label{tabimage}%
\begin{tabular}{@{}llllll@{}}
\toprule
Methods & 40X  & 100X & 200X & 400X & Mean\\
\midrule
pdMISVM    & $87.92\pm0.9$  & $89.04\pm2.4$ & $88.91\pm2.1$ & $85.78\pm1.9$ & $87.91\pm1.8$ \\
DenseNet121    & $\bm{94.26\pm3.2}$   & $92.71\pm0.4$ & $83.90\pm2.8$ & $82.74\pm1.5$ & $88.40\pm2.0$ \\
ResHist-Aug  & $ 87.40\pm3.0$  & $87.26\pm3.5$ & $91.15\pm2.3$ & $86.27\pm2.2$ & $88.02\pm2.8$ \\
MPCS-OP    & $93.26\pm3.4$ & $93.45\pm2.9$ & $92.45\pm3.8$ & $89.57\pm3.0$ & $92.18\pm3.3$  \\
MPCS-RP    & $92.72\pm3.9$ & $92.72\pm4.0$ & $91.91\pm3.2$ & $88.56\pm3.9$ & $91.48\pm3.7$\\
\textcolor{black}{GLNET} & $90.01\pm0.0$ & $92.32\pm0.0$ & $91.98\pm0.0$ & $91.51\pm0.0$ & $91.45\pm0.9$ \\
\textcolor{black}{MSB}    & $93.76\pm2.3$ & $92.04\pm3.4$ & $91.31\pm2.9$ & $89.81\pm2.2$ & $91.73\pm2.7$ \\
\textcolor{black}{MIB}    & $90.14\pm1.2$ & $91.83\pm1.4$ & $91.45\pm1.6$ & $89.90\pm1.9$ & $90.83\pm1.5$ \\
Ours    & $93.00\pm2.8$ & $\bm{93.51\pm2.1}$ & $\bm{95.28\pm1.9}$ & $\bm{92.73\pm1.8}$ & $\bm{93.63\pm2.2}$ \\
\botrule
\end{tabular}
\end{table}

\begin{table}[t]
\caption{\textcolor{black}{Comparison of patient-level accuracy with previous methods for four magnification factors of the BreakHis dataset.}}\label{tabpatient}%
\begin{tabular}{@{}llllll@{}}
\toprule
Methods & 40X  & 100X & 200X & 400X & Mean\\
\midrule
PFTAS    & $82.03\pm2.8$ & $81.02\pm3.9$ & $83.58\pm3.1$ & $81.17\pm4.4$ & $81.95\pm3.6$ \\
DenseNet121    & $92.02\pm0.9$ & $90.21\pm2.4$ & $81.94\pm1.7$ & $80.09\pm0.7$ & $86.06\pm1.4$ \\
ResHist-Aug & $87.47\pm3.2$   & $ 88.15\pm3.0$ & $92.52\pm2.8$ & $87.78\pm2.5$ & $88.98\pm2.9$ \\
MPCS-OP    & $93.00\pm3.7$ & $93.26\pm3.1$ & $92.28\pm2.9$ & $88.74\pm3.9$ & $91.82\pm3.3$  \\
MPCS-RP    & $92.72\pm3.5$ & $93.57\pm2.9$ & $92.23\pm3.9$ & $88.40\pm3.1$ & $91.73\pm3.3$ \\
\textcolor{black}{MIB}    & $92.30\pm1.5$ & $93.00\pm1.1$ & $92.50\pm1.4$ & $89.60\pm2.0$ & $91.85\pm1.5$ \\
\textcolor{black}{quadtree16}    & $\bm{94.06\pm0.1}$ & $\bm{94.50\pm0.3}$ & $92.03\pm0.6$ & $91.16\pm0.2$ & $92.93\pm1.4$ \\
Ours    & $93.99\pm3.0
$ & $91.34\pm6.6$ & $\bm{94.80\pm2.4}$ & $\bm{92.84\pm2.9}$ & $\bm{93.24\pm3.7}$ \\
\botrule
\end{tabular}
\end{table}

\subsection{Results}
In this section, we compare the results of the proposed method with those of the methods, PFTAS \cite{7312934}, pdMISVM \cite{seo2022scaling}, MPCS-OP \cite{chhipa2023magnification}, MPCS-RP \cite{chhipa2023magnification}, DenseNet121 \cite{man2020classification}, ResHist-Aug \cite{gour2020residual}, \textcolor{black}{MSB\cite{taheri2023magnification}, MIB\cite{taheri2024comprehensive}, GLNET\cite{khan2024glnet}, and quadtree16\cite{xiao2024convolutional}.} Table \ref{tabimage} and Table \ref{tabpatient} illustrate the image-level and patient-level accuracy achieved by different methods, respectively.

As shown in Table \ref{tabimage}, the model proposed in this study has outperformed the existing methods in terms of accuracy across three magnification factors while exhibiting a slight decrease in the other magnification factor. On average, this method has demonstrated an improvement of $1.45\%$ in the image level, and $0.31\%$ in the patient level accuracy compared to the previous models.

The accuracy metric is unsuitable for evaluating this dataset due to the unequal distribution of data points across categories. This can introduce bias and lead to a false high accuracy. To overcome this issue, We adopted weighted metrics such as balanced accuracy and weighted F1 metric, which assign weights to each category based on the number of samples available.

To evaluate the generalization capability of the trained model, we utilized the weights of models that were previously trained on each of the four magnification factors of the BreakHis dataset. These weights served as the initial starting point for training the BACH dataset. As shown in Table \ref{tabbach}, the model can achieve an accuracy above 90\% after only 20 epochs, indicating that this method has successfully learned discriminative features specific to histopathology images, and can generalize well to other histopathology datasets with minimal additional training. Moreover, a decreasing trend in the accuracy can be observed as the magnification factor increases which suggests that the BACH dataset images are closer to the 40X magnification factor of the BreakHis. This observation is further supported by visually inspecting the two datasets images.

\subsubsection{\textcolor{black}{Ablation Study}}
Eq. \ref{floss-def}, presented in this method, is derived from a combination of three loss functions: self-supervised loss \cite{chen2020simple}, supervised contrastive loss \cite{khosla2020supervised}, and elimination loss \cite{huynh2022boosting}. The specific formulas for each of these loss functions are described below:

\begin{equation}
\ell_{self(i)} = -log (\frac{exp(z_i \cdot z_j/\tau)}{\sum_{k=1}^{2N} \mathbbm{1}_{k \ne i} exp(z_i \cdot z_k/\tau)})
\label{sim-def}
\end{equation}

\begin{equation}
\ell_{sup(i)} = \frac{-1}{|P(i)|} \sum_{p\in P(i)} log(\frac{exp(z_i \cdot z_p/\tau)}{\sum_{k=1}^{2N} \mathbbm{1}_{k \ne i} exp(z_i \cdot z_a/\tau))})
\label{sup-def}
\end{equation}

\begin{equation}
\ell _{elim(i)} = -log (\frac{exp(z_i \cdot z_j/\tau)}{\sum_{k=1}^{2N} \mathbbm{1}_{[k \ne i, k \notin P(i)]} exp(z_i \cdot z_k/\tau)}).   \label{eli-def}
\end{equation}

The structure of our presented loss function is similar to Eq. \ref{sup-def}. By introducing a coefficient $\lambda$, and the $Q$ set, we have incorporated the effect of the Eq. \ref{eli-def}. Additionally, by utilizing augmentations for creating a positive pair for images, we have also integrated the impact of the Eq. \ref{sim-def} into our approach.

\begin{table}[t]
\caption{Examining the impact of each term in the combination of introduced loss functions.}\label{tabloss}%
%\setlength{\tabcolsep}{0.4\tabcolsep}
%\begin{center}
\begin{tabular}{|c|c|c|c|c|c|c|c|}
%\hline
\toprule
Method & Comb1  & Comb2 & Comb3 & Comb4 & Comb5 & Comb6 & Comb7\\
%\hline
\midrule
Sup    & \checkmark &  &   & \checkmark  & \checkmark  &  & \checkmark  \\
Elim    &  & \checkmark  &  & \checkmark  &  & \checkmark  & \checkmark  \\
Self    &  &  & \checkmark  &  & \checkmark  & \checkmark  & \checkmark  \\
Accuracy    & 93.68 & 89.50 & 91.16 & 95.08 & 93.86 & 91.35 & 96.74 \\
%\hline
\botrule
\end{tabular}
%\end{center}
\end{table}

Table \ref{tabloss} analyses the impact of each of the loss functions \ref{sim-def}, \ref{sup-def}, and \ref{eli-def} and their combinations. These experiments have only been conducted on one fold, with the primary objective being the comparison of various loss functions.%, rather than a comparison with previous methods.
The results displayed in the table indicate that $\ell_{sup}$ has the most significant effect in improving the outcomes. This can be due to the fact that $\ell_{sup}$ leverages the labels to eliminate the impact of false negatives in the numerator. Moreover, this loss function aims to minimize the similarity between the anchor and both negative and positive pairs within a batch in the denominator. By doing so, $\ell_{sup}$ effectively prevents the network from being disrupted in the presence of a not similar positive pair. Another aspect of this loss function is the selection of positive pairs, which is done among all patients with similar diseases. By bringing the representations of images from different patients closer together, we enable the network to generalize its learning to the images of unseen patients in the test set.

The second most influential loss function is $\ell_{elim}$. In this loss function, the positive pair is generated using data augmentation techniques, while the negative pairs are selected from images with the opposite label compared to the anchor. In essence, this loss function serves as a safety net, ensuring that only pairs we have complete confidence in being either similar or non-similar are brought closer together or separated.

The $\ell_{self}$ has the most negligible impact among all the losses. This loss is similar to $\ell_{sup}$, with the only difference being the positive pairs. Including this term increases the influence of the positive pair generated through data augmentation, which in turn leads to an improvement in accuracy.

\section{Discussion}

The results from previous methods demonstrate that as the magnification factor increases, the accuracy decreases. This trend can be attributed to higher magnification images capturing more intricate and internal details of the cells, in contrast to lower magnification images that primarily capture the cell structure and overall patterns. Consequently, the higher magnification images pose a more significant challenge for neural networks to learn and classify accurately. However, this is not the case in the proposed model as we were able to generally achieve a high accuracy in all of the four magnification factors. This can be attributed to the characteristic of the proposed loss function. The combination of self-supervised, supervised contrastive, and elimination ensures that the model learns a comprehensive set of features. Self-supervised learning helps capture a wide range of visual patterns, while supervised contrastive loss refines these features based on class labels. Elimination part of loss further ensures that only reliable features are considered. This holistic approach helps the model to effectively capture and utilize detailed features present in high magnification images as well as the low magnifications. This uniformly high performance across all magnification levels is the main reason our overall accuracy generally surpasses that of the models presented in previous cited experiments.

\begin{table}[t]
\caption{Computed score for image-level classification on the BACH dataset using the weights trained on the four magnification factors of the BreakHis dataset.}\label{tabbach}%
% \setlength{\tabcolsep}{0.4\tabcolsep}
%\begin{center}
\begin{tabular}{@{}llllllll@{}}
\toprule
%\hline
Mag & Precision  & Recall & Weight-F1 & Acc & Balance-Acc & Kappa & Dice\\
%\hline
\midrule
40X    & 92.61  & 92.50  & 92.49 & 92.50 & 92.50 & 85.00 & 92.31 \\
100X    & 91.28 & 91.25 & 91.25 & 91.25 & 91.25 & 82.50 & 91.14 \\
200X    & 90.40 & 90.00 & 89.97 & 90.00 & 89.99 & 80.00 & 89.47  \\
400X    & 90.40 & 89.99 & 89.96  & 90.00 & 90.00 & 80.00 & 89.48 \\
%\hline
\botrule
\end{tabular}
%\end{center}
\end{table}

\section{Conclusion and Future Work}
The proposed approach aims to enhance the binary classification of the BreakHis dataset by extending the supervised contrastive learning method to reduce the number of false positives in addition to false negatives. Furthermore, various augmentations specific to the histopathology domain were employed to improve the model's performance and generalization across different H\&E stains. Consequently, the model exhibited superior performance compared to prior methods in three magnification factors while also achieving higher accuracy on average across all four magnification factors.

In the proposed method, a potential concern arises from considering all images with the same label as positive pairs during the initial representation stage. This can be problematic since cancerous images can exhibit significant variations, even within a single patient. In this approach, all images, including those from the same patient and different ones, are treated as positive pairs. Consequently, some pairs which were considered positive may not exhibit sufficient similarity, and actually present a challenge for the network. The loss function, in attempting to increase their similarities, can lead to a deviation from the model's ultimate goal. To overcome this issue, our future work will aim to extract pairs more intelligently, beyond relying solely on the labels. 

\section{Statement and Declaration}

\subsection{Funding}
The authors declare that no funds, grants, or other support were received during the preparation of this manuscript.
\subsection{Competing Interests}
The authors have no relevant financial or non-financial interests to disclose.
\subsection{Author Contributions}
All authors contributed to the study conception and design. Material preparation, data collection, and analysis were performed by all authors. The first draft of the manuscript was written by the first auther and all authors commented on several versions of the manuscript. All authors read and approved the final manuscript.%Matina Mahdizadeh Sani, Ali Royat, and Mahdieh Soleymani Baghshah. The first draft of the manuscript was written by Matina Mahdizadeh Sani and all authors commented on previous versions of the manuscript. All authors read and approved the final manuscript.
\subsection{Ethics Approval}
The datasets used in this study consist of histopathology images from breast cancer tissues. These images have been obtained from prior publications and are being reused for the current research.

\bibliography{sn-bibliography}% common bib file

\section*{Biography}

\subsection*{Matina Mahdizadeh Sani}
Matina Mahdizadeh Sani is currently a graduate student at University of Waterloo, Canada. She got her bachelor’s degree from Sharif university of technology, Tehran, Iran in 2023. Her field of study is machine learning and computer vision.

\subsection*{Ali Royat}
Ali Royat is a graduate student (B.Sc. and M.Sc.) in Electrical Engineering at the Sharif University of Technology, Tehran, Iran. His expertise lies in the application of deep learning to Computer Vision and signal processing.

\subsection*{Mahdieh Soleymani Baghshah}
Mahdieh Soleymani received her BS, MS, and PhD degrees from the Department of Computer
Engineering, Sharif University of Technology, Iran, in 2003, 2005, and 2010, respectively. Her MS
and PhD theses were in the field of Machine Learning. She joined Sharif University of Technology as
an assistant professor of Computer Engineering and founded the Machine Learning Laboratory
(MLL) in 2012. She is currently an associate professor at Sharif University of Technology and
the director of MLL. Her research interests include machine learning and deep learning. She has
worked on various types of data including images, texts, videos, omics, and financial time series.

\end{document}